\documentclass[a4paper, 10 pt, conference]{IEEEtran}
\IEEEoverridecommandlockouts
\usepackage[hmargin=40pt,vmargin=94pt]{geometry}

\usepackage{cite}
\usepackage{amsmath,amssymb,amsfonts}
\usepackage{algorithmic}
\usepackage{graphicx}
\usepackage{textcomp}

\usepackage{tabularx}
\usepackage{multirow}
\usepackage{booktabs}

\usepackage{xcolor}
\usepackage{siunitx}
\usepackage{graphicx}
\graphicspath{{figures}}
\def\BibTeX{{\rm B\kern-.05em{\sc i\kern-.025em b}\kern-.08em
    T\kern-.1667em\lower.7ex\hbox{E}\kern-.125emX}}

\begin{document}

\title{Design of a Double-joint Robotic Fish Using a Composite Linkage}
\makeatletter
\newcommand{\linebreakand}{%
\end{@IEEEauthorhalign}
\hfill\mbox{}\par
\mbox{}\hfill\begin{@IEEEauthorhalign}
}
\makeatother
\author{\IEEEauthorblockN{Ruijia Zhang}
\IEEEauthorblockA{\textit{School of Power and Mechanical Engineering} \\
\textit{Wuhan University}\\
Wuhan, China \\
2020302191853@whu.edu.cn}\\
\IEEEauthorblockN{Min Li$^{*}$}
\IEEEauthorblockA{\textit{$^{*}$Corresponding author}\\
	\textit{School of Power and Mechanical Engineering} \\
	\textit{Wuhan University}\\
	Wuhan, China \\
	00032960@whu.edu.cn}
\and	
\IEEEauthorblockN{Wenke Zhou}
\IEEEauthorblockA{\textit{School of Mechanical Science and Engineering} \\
\textit{Huazhong University of Science and Technology}\\
Wuhan, China \\
m202370637@hust.edu.cn}\\
\IEEEauthorblockN{Miao Li$^{*}$}
\IEEEauthorblockA{\textit{$^{*}$Corresponding author}\\
\textit{the Institute of Technological Sciences} \\
\textit{Wuhan University}\\
Wuhan, China \\
miao.li@whu.edu.cn}
}
\maketitle
\begin{abstract}
	Robotic fish is one of the most promising directions of the new generation of underwater vehicles. Traditional biomimetic fish often mimic fish joints using tandem components like servos, which leads to increased volume, weight and control complexity. In this paper, a new double-joint robotic fish using a composite linkage was designed, where the propulsion mechanism transforms the single-degree-of-freedom rotation of the motor into a double-degree-of-freedom coupled motion, namely caudal peduncle translation and caudal fin rotation. Motion analysis of the propulsion mechanism demonstrates its ability to closely emulate the undulating movement observed in carangiform fish. Experimental results further validate the feasibility of the proposed propulsion mechanism. To improve propulsion efficiency, an analysis is conducted to explore the influence of swing angle amplitude and swing frequency on the swimming speed of the robotic fish. This examination establishes a practical foundation for future research on such robotic fish systems.
\end{abstract}
\section{Introduction}
	Underwater vehicles currently play a pivotal role in ocean exploration. Traditional propeller-based propulsion systems are considered unsuitable for practical applications due to their inefficiency, limited maneuverability, and lack of robustness to environmental disturbance. Consequently, significant attention has been placed on biomimetic underwater robots, drawing inspiration from the distinctive propulsion patterns exhibited by aquatic organisms \cite{ref1}. Fish, as representative aquatic organisms, offer abundant inspiration for the design of biomimetic underwater robots due to their ubiquity and ease of observation \cite{ref2}. Fish movement is broadly categorized into Body-Caudal Fin (BCF) propulsion and Median-Pectoral Fin (MPF) propulsion \cite{ref3}. BCF swimming, characterized by coordinated body and caudal fin swing, is favored for its advantages in speed, efficiency, and rapid start-up. As a result, the majority of research in biomimetic robotic fish focuses on BCF swimming mode.
	
	Following the development of MIT's ``Robo tuna'' \cite{ref4}, various research institutions have been actively creating prototypes of biomimetic robotic fish. The majority of robotic fish utilize rigid motors due to their ease of implementation and convenience compared to flexible-driven robots \cite{ref5}. Among the rigid motor-driven robotic fishes, there are primarily two types: single-joint and multi-joint drives. Robotic fishes with single-joint drives exhibit simpler structures, smaller sizes, and higher Strouhal numbers. Examples include the ostraciiform robotic fish developed by Daisy Lachat et al., featuring a motor-driven gear mechanism \cite{ref6}, and the innovative bionic fish with a double-cam mechanism studied by Song et al. \cite{ref7}.
	
	In comparison to single-joint driven robotic fish, those with multiple joints offer increased flexibility and the capacity to carry numerous functional sensors. Consequently, this study focuses on the design and exploration of bionic fish driven by multiple joints. Currently, several multi-joint driven bionic fish have been developed. For example, Nakashima et al. designed a self-propelled, two-joint robotic dolphin with a crank and rocker mechanism, achieving a notable maximum speed of 1.15 m/s \cite{ref8,ref9}. The bionic underwater robot developed by Cheng et al. utilizes two rigid servo motors and an interactive gear system, enabling a range of intricate movements \cite{ref10}. Liang et al. have developed a series of thunniform robotic fish, driven by two servo motors that control two parallel joints to simulate the coupled motion of the fish tail \cite{ref11,ref12}. 
	
	To sum up, the majority of multi-joint fish robots are driven by motors either directly connected serially or indirectly via a transmission mechanism. However, connecting motors serially for joint swing motions deviates from the authentic swimming posture of fish, resulting in reduced propulsion efficiency \cite{ref13}. 
	
	\newgeometry{hmargin=40pt,vmargin=68pt}
	On the other hand, the use of transmission mechanisms to emulate the coupled motion of fish tail, whether driven by multiple motors or compliant passive drives (such as springs), increases the difficulty in adhering to input control patterns and also leads to increased volume and weight \cite{ref8,ref11}.
 
	Therefore, this paper proposed a double-joint biomimetic fish using a composite linkage mechanism, offering a simpler and more reliable mechanical structure. This design enables dual-joint coupled motion of the caudal fin and caudal peduncle through a single motor drive, closely aligning with real fish movement. Moreover, the motion parameters of the robotic fish play a key role in propulsion efficiency \cite{ref14}. Thus, following the completion of the structural design, a prototype is manufactured, and experiments are conducted to investigate the impact of motion parameters, specifically the swing angle amplitude and swing frequency, on the swimming speed of the robotic fish, aiming to enhance propulsion efficiency.
\section{Principle and Design of Propulsion Mechanism}

\subsection{BCF Mode in Fish Locomotion}
	For fish utilizing the BCF swimming mode, such as those in the Carangidae family, the anterior half of the body remains nearly stationary during the swimming process. Meanwhile, the posterior third of the body undergoes oscillations, thereby generating forward thrust. According to observations from experiments conducted by Videler and Hess \cite{ref15,ref16}, the undulating motion of the fish's body can be represented by the following equation:
\begin{equation} \label{eq1}
	y(x,t)=A(x)sin(kx+\omega t) \tag{1}
\end{equation} 
	Where $x$ is the coordinate of the fish body's midline along the central axis of the fish's body, with the head position taken as 0, and $y$ is defined as the relative distance from the horizontal axis to the midline of fish body; $k$ is the body wave number ($k=2\pi/\lambda$, $\lambda$ is the body wavelength); $\omega$ is the body wave frequency ($\omega=2\pi/T$, $T$ is the period of the fish body wave); $A(x)$ represents the amplitude envelope of the fish body wave curve, and its expression is given by:
\begin{equation} \label{eq2}
	A(x)=c_1+c_2x+c_3x^2 \tag{2}
\end{equation} 

	Here, $c_1$ is the lateral amplitude of the geometrical center of the fish's swimming, $c_2$ is the linear wave amplitude envelope, $c_3$ is the quadratic wave amplitude envelope.
\begin{table}[b!]
	\caption{Parameters of Fish Body Wave Equation}
	\label{tab1}
	\setlength{\tabcolsep}{5mm}
	\begin{center}
		\begin{tabular}{ccccc}
			\toprule
			$c_1$   & $c_2$   & $c_3$   & $\lambda (m)$  & $\omega (rad/s)$ \\ \midrule
			0.02 & 0.08 & 0.16 & 0.95 & $2\pi$    \\ \bottomrule
		\end{tabular}
	\end{center}
\end{table}	

	The schematic representation of the fish body wave during one motion period is depicted in Fig.~\ref{figl}. The red curve represents the midline of the fish body, as described by \eqref{eq1}. The corresponding parameter values, derived from experimental data provided by Videler and Hess \cite{ref15}, are detailed in Table \ref{tab1}. Additionally, Fig.~\ref{fig2} presents the fish body wave curve at various time points along with its amplitude envelope curve, derived from the aforementioned experimental data.
	
	From Fig.~\ref{fig2}, it is evident that the swing amplitude progressively increases from the fish's head to its tail. Therefore, fish belonging to the Carangidae family primarily rely on the tail to generate propulsive force. The tail's motion can be characterized as a composite movement involving the lateral translation of the caudal peduncle and the swing of the caudal fin, as depicted in Fig.~\ref{figl}. According to the wave equation, the lateral translation $H(t)$ and swing angle $\theta(t)$ of the tail's motion can be articulated through the following equations:
\begin{equation} \label{eq3}
	H(t)=H_{max}sin(\omega t) \tag{3}
\end{equation} 
\begin{equation} \label{eq4}
	\theta(t)=\theta_{max}sin(\omega t+\varphi) \tag{4}
\end{equation}
Where $H_{max}$ represents the limit of lateral translation, $\theta_{max}$ represents the limit of swing angle; $\omega$ is the swing frequency; $\varphi$ denotes the phase difference between lateral translation and swing angle.	
\begin{figure}[t!]
	\centering
	\includegraphics[width=\linewidth]{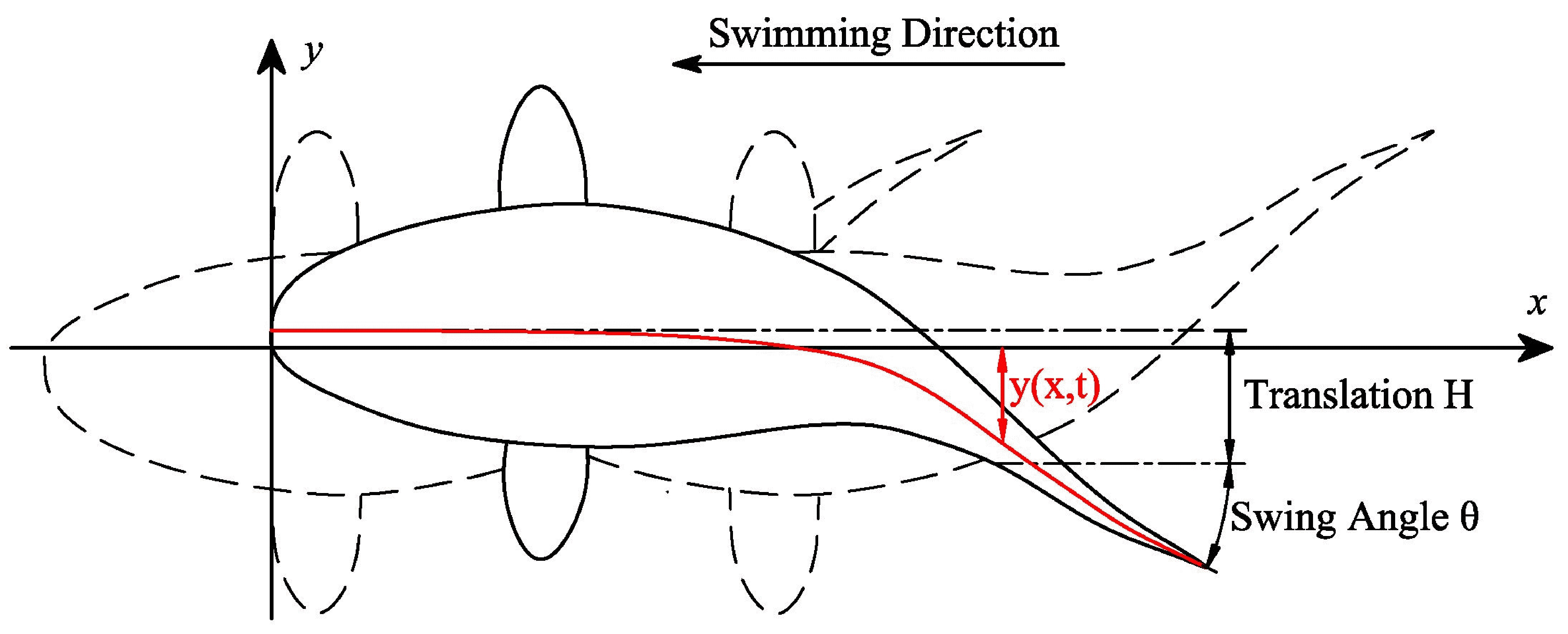}
	\caption{Schematic diagram of the fish body wave during one motion period. The red curve represents the midline of the fish body.}
	\label{figl}
\end{figure}
\begin{figure}[t!]
	\centering
	\includegraphics[width=\linewidth]{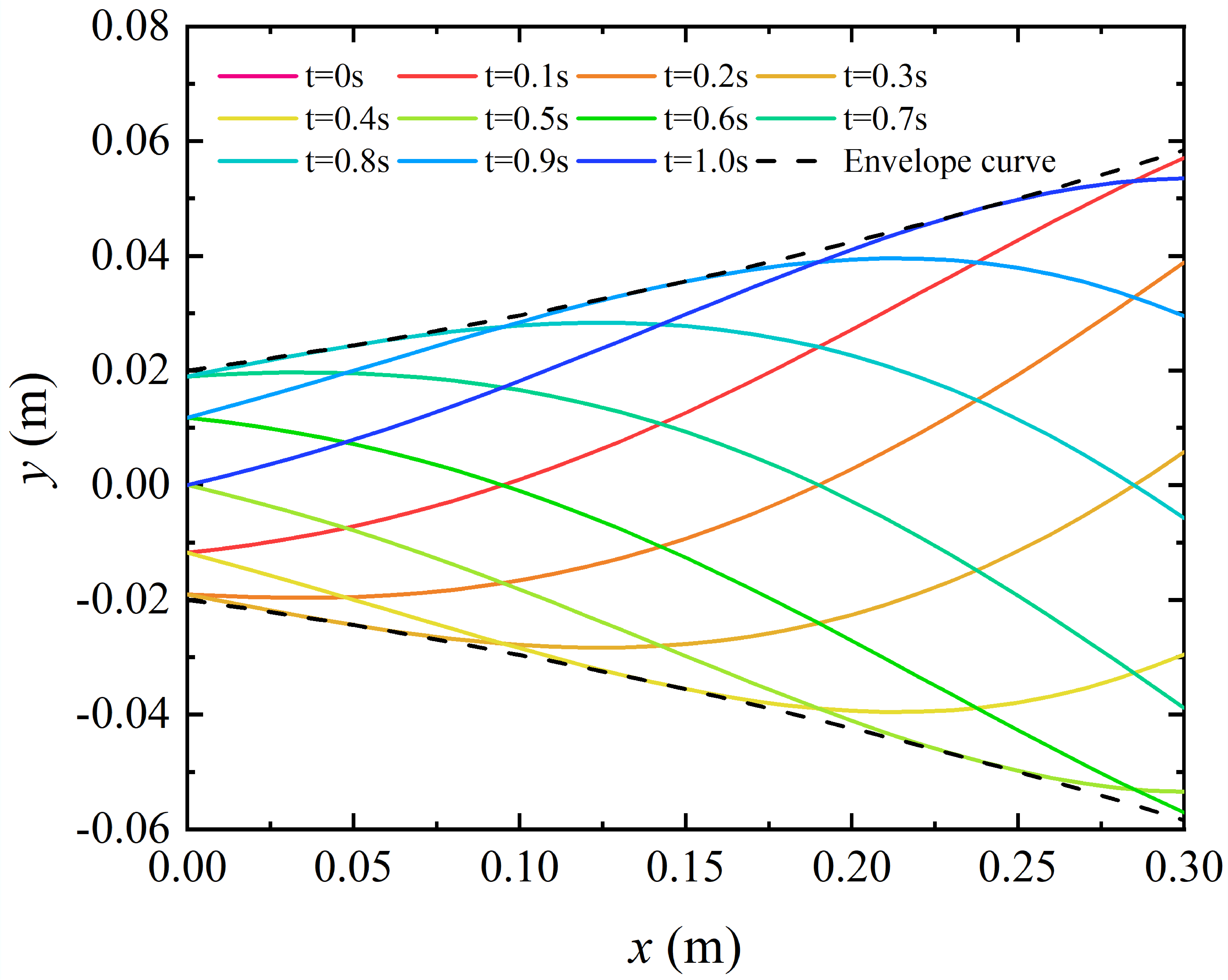}
	\caption{Fish body wave curve at various time points along with its amplitude envelope curve.}
	\label{fig2}
\end{figure}
\subsection{Propulsion Mechanism Principle}
	The movement of the robot's fishtail is typically achieved by connecting multiple motors in series to control the rotation of individual joints separately \cite{ref13}. However, in this study, a unique composite linkage mechanism was designed to fit the coupled motion trajectory of the fishtail, as depicted in Fig.~\ref{fig3} (a). This mechanism exhibits a reliable structure, simple control, and efficient propulsion.

	The transmission diagram of the propulsion mechanism is illustrated in Fig.~\ref{fig3} (b). At one end, two cranks, set at a fixed angle, are connected at point O and rigidly fixed to the output shaft of the motor. At the opposing end, the fixed pins, designated as M and N, are designed to slide smoothly within the chutes of two linkages parallel to the \textit{Y}-axis, respectively. Each linkage is attached to a slider, R and S, at one end, allowing translational motion along the \textit{Y}-axis within the sliding rail. At the opposing end, the linkages articulate with two swing arms, denoted as AC and BC. Both swing arms pivot at point C, with the caudal fin attached to the swing arm AC at the same pivotal point, C. 
\begin{figure}[t!]
	\centering
	\includegraphics[width=\linewidth]{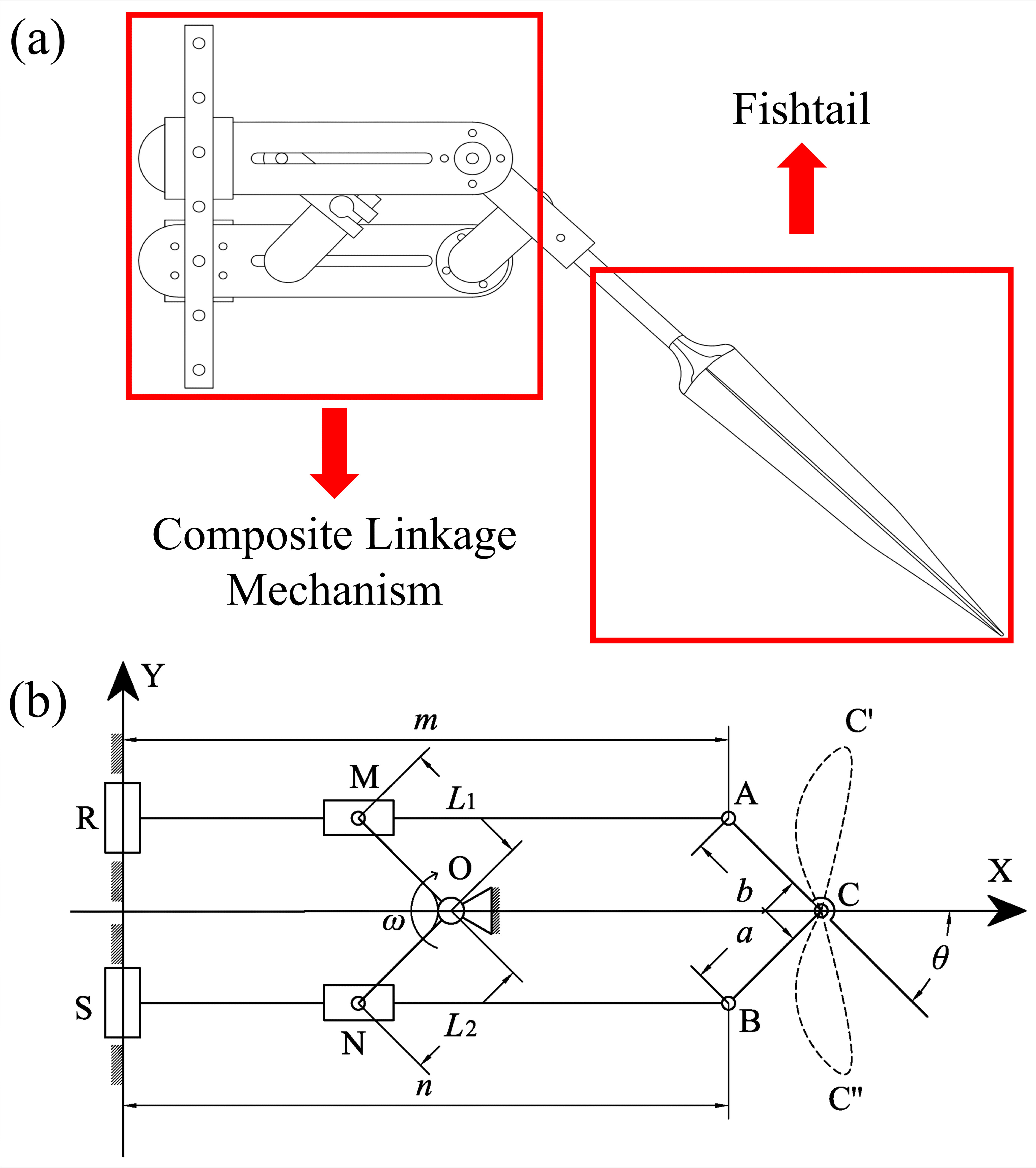}
	\caption{(a) Schematic diagram of the propulsion mechanism, including the composite linkage mechanism and the fishtail. (b) Transmission diagram of the propulsion mechanism.}
	\label{fig3}
\end{figure}	
	
	Within one motion period, the motor rotates unidirectionally, driving the two cranks to rotate around the pivotal point O. This rotational motion induces the linkages to undergo translational movement along the \textit{Y}-axis. The point C traces a trajectory through points C-C'-C-C''-C, forming an 8-shaped path that corresponds to the swimming pattern observed in fish of the Carangidae family.

\subsection{Motion Analysis}\label{AA}
	During straight-line cruising, the fishtail swing is symmetric, indicating equal lengths of the two cranks, i.e., $L_1$ = $L_2$. The lengths of the two linkages are denoted as $m$ and $n$, while the length of the driving swing arm AC and driven swing arm BC is represented by $a$ and $b$, respectively. Due to the special boundary conditions where the two linkages and the two swing arms coincide during motion of this mechanism, the following conditions should be satisfied:
\begin{equation} \label{eq5}
	m+b=n+a \tag{5}
\end{equation}

	When the cranks rotate with an angular velocity $\omega$, the translation of points A and B in the \textit{Y}-direction are given by:
\begin{equation} \label{eq6}
	\left\{  
	\begin{array}{lr}  
		S_A = L_1 cos(\omega t), &  \\  
		S_B = L_2 (\omega t+\varphi), &   
	\end{array}  
	\right.
	\tag{6}
\end{equation} 
	Where $\varphi$ is the angle between the two cranks. Based on the geometric relationships of the mechanism, the motion equations for the mechanism are:
\begin{equation} \label{eq7}  
	\left\{  
	\begin{array}{lr}  
		L_1 = L_2 &  \\  
		a cos\theta = b cos\theta +m-n &  \\   
		S_A-S_B=a sin\theta+b sin\theta &  \\
		S_{CY} = S_A -b sin \theta &  \\
		S_{CX} = b cos \theta &
	\end{array}  
	\right.
	\tag{7}
\end{equation} 
	Here, $\theta$ is the swing angle, $S_{CY}$ is the displacement of point C along the \textit{Y}-direction, and $S_{CX}$ is the displacement of point C along the \textit{X}-direction. Therefore, the derivation of the swing angle $\theta$ and the displacement $S_{CY}$ is as follows:
\begin{equation} \label{eq8} 
	\left\{  
	\begin{array}{lr}  
		\theta=\arcsin\displaystyle{\frac{\Delta S^2+b^2+\delta^2-a^2}{2b\sqrt{\Delta S^2+\delta^2}}}+\arctan\displaystyle{\frac{\delta}{\Delta S}} &  \\   
		[2ex]S_{CY}=S_A-b\sin(\arcsin\displaystyle{\frac{\Delta S^2+b^2+\delta^2-a^2}{2b\sqrt{\Delta S^2+\delta^2}}}\\\\+\displaystyle{\arctan\frac{\delta}{\Delta S})} &
	\end{array}  
	\right.
	\tag{8}
\end{equation} 
	Where $\Delta S = S_A - S_B$, $\Delta = m - n$. This motion equation still differs from \eqref{eq3} and \eqref{eq4}. To better fit the real fishtail motion trajectory, an additional constraint of $m = n$ is introduced, leading to $a = b$. In this case, the motion equations for the linkages become:
\begin{equation} \label{eq9} 
	\left\{  
	\begin{array}{lr}  
		\theta=\arcsin\displaystyle{\frac{\Delta S^2+b^2-a^2}{2b\Delta S}} &  \\   
		[2ex]S_{CY}=S_A-\displaystyle{\frac{\Delta S^2+b^2-a^2}{2\Delta S}} &
	\end{array}  
	\right.
	\tag{9}
\end{equation} 	
Substituting \eqref{eq6} into \eqref{eq9} results in:
\begin{equation} \label{eq10}  
	\left\{  
	\begin{array}{lr}  
		\theta=\arcsin(\displaystyle{\frac{L_1}{\mathrm{a}}}\sin(\omega t+\displaystyle{\frac{\varphi}{2})}\sin(\displaystyle{\frac{\varphi}{2}))} &  \\   
		[2ex]S_{CY}=\displaystyle{\frac{L_1}{2}}\left(\cos(\omega t+\varphi)+\cos(\omega t)\right) &
	\end{array}  
	\right.
	\tag{10}
\end{equation} 

	Referring to the real fish motion parameters, i.e., the maximum swing angle $\theta_{max} = 43.88^{\circ}$, and the maximum translation of the caudal peduncle $H_{max} = \SI{15.81}{\milli\metre}$, when the phase difference $\varphi$ is $90^{\circ}$, the values of structural parameters are listed in Table \ref{tab2}.		
\begin{table}[b!]
	\caption{Mechanism Parameters of Robotic Fish}
	\label{tab2}
	\setlength{\tabcolsep}{4mm}
	\begin{center}
		\begin{tabular}{ccccc}
			\toprule
			$L_1 (mm)$  & $L_2 (mm)$   & $\varphi^{\circ}$   & $a (mm)$  & $b (mm)$ \\ \midrule
			22.36 & 22.36 & 90 & 22.81 & 22.81    \\ \bottomrule
		\end{tabular}
	\end{center}
\end{table}
	
	According to the analysis, when $\sin(\omega t+\frac{\varphi}{2})=1$, the relationship between the maximum swing angle and the structural parameters is:
\begin{equation} \label{eq11}
	\theta_{\max}=\arcsin(\displaystyle{\frac{L_{1}}{a}}\sin{(\displaystyle{\frac{\varphi}{2}})}) \tag{11}
\end{equation}
\begin{figure}[t!]
	\centering
	\includegraphics[width=\linewidth]{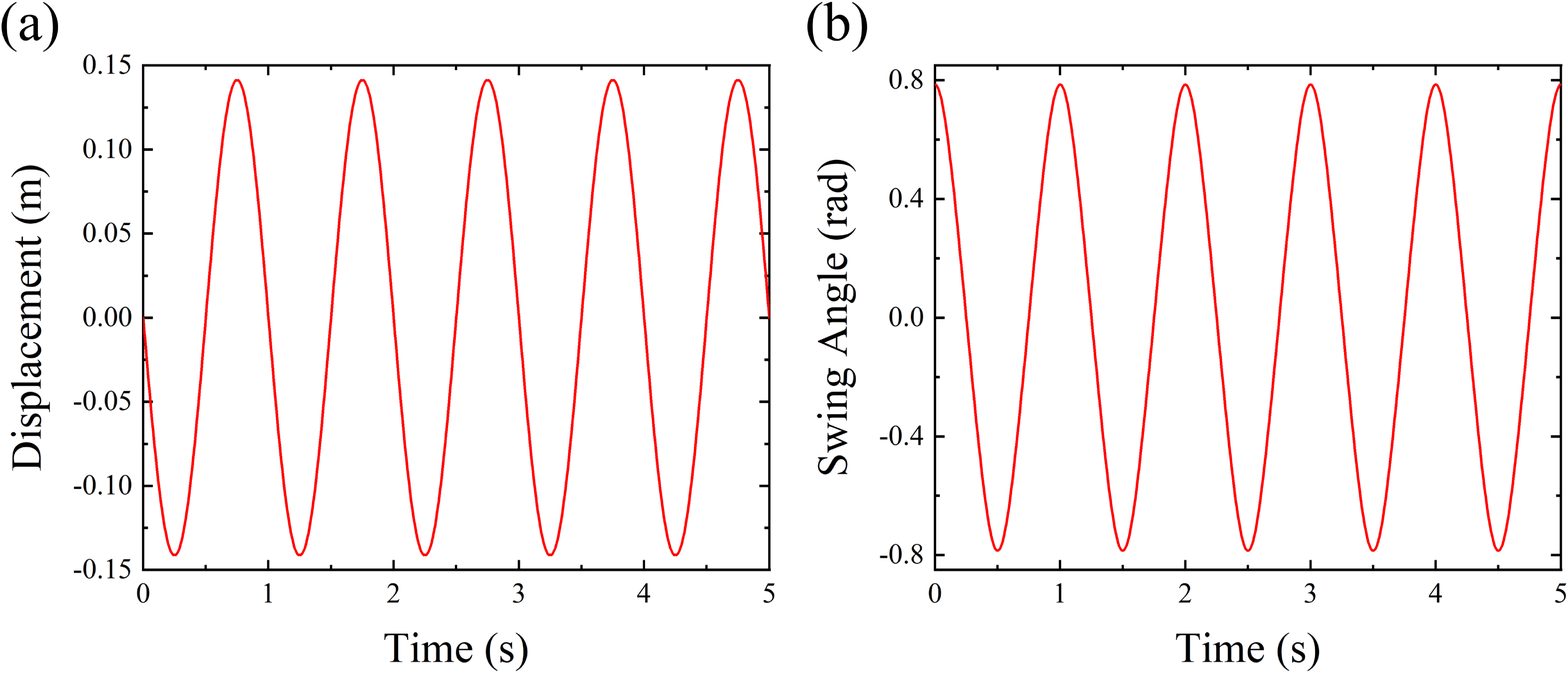}
	\caption{(a) Motion pattern of lateral displacement with time. (b) Motion pattern of swing angle with time.}
	\label{fig4}
\end{figure}

	Substituting the above parameters into \eqref{eq10}, the motion patterns of lateral displacement $S_{CY}$ and swing angle $\theta$ with time are shown in Fig.~\ref{fig4}. The results indicate a close resemblance to the sinusoidal motion described by \eqref{eq1}.
\section{Prototype Experiments and Analysis}
\subsection{Overall Structural Design and Prototyping}
	The robotic fish adopts a modular design, primarily comprising head shell, driving module, fishtail skeleton, flutter mechanism, pitching mechanism, and control module, as depicted in Fig.~\ref{fig9}. Building upon the previously designed propulsion mechanism, the driving module utilizes a \SI{24}{\volt} double-shaft DC motor to drive the cranks, with a steering servo to control the fish body's direction. An externally designed fishtail skeleton is employed to simulate a biological fishtail. The pitching mechanism, located beneath the steering servo, consists of a lead screw with added weight blocks to regulate the fish's pitch angle. The control module and flutter mechanism are situated within the fish head shell. A Raspberry Pi functions as the main control board, interfacing with various motor drive modules and sensor control modules. It controls the DC motor of the driving module and coordinates the operation of three additional servos for different locomotion modes, including straight swimming, steering, and pitching.
	
	To ensure rigidity in the driving module, aluminum alloy is chosen for the rods. For underwater reliability of the electrical components in the control module, a waterproof box made of aluminum alloy seals the control module. Other parts are 3D printed using epoxy resin material.
	In the design of the robotic fish, dimensions are proportionally enlarged according to the structural dimensions of live fish, as illustrated in Fig.~\ref{fig10}. The specified dimensions are set as follows: a fish head length of \SI{255}{\milli\metre} and a fishtail length of \SI{285}{\milli\metre}.	

\subsection{Experimental Methods}
	The experimental platform comprises three main components: a water tank, a camera bracket, and a high-speed camera. The water tank measures $\SI{3}{\metre} \times \SI{2}{\metre} \times \SI{0.6}{\metre}$. A bracket spanning the water tank suspends the camera above the robotic fish's swimming path. The velocity estimation method involves capturing images of the robotic fish in a stable swimming state from the video, where the motion direction remains constant and speed fluctuations are minimal. The distance traveled by the fish is then determined from these images, allowing for the calculation of the prototype robotic fish's swimming speed. The experiment investigates the swimming speed of the robotic fish under different tail fin swing frequencies and swing angle amplitudes. Fig.~\ref{fig11} displays images captured from the camera, showcasing successive states of the robotic fish's motion at quarter-cycle intervals during one swimming cycle.
\begin{figure}[t!]
	\centering
	\includegraphics[width=\linewidth]{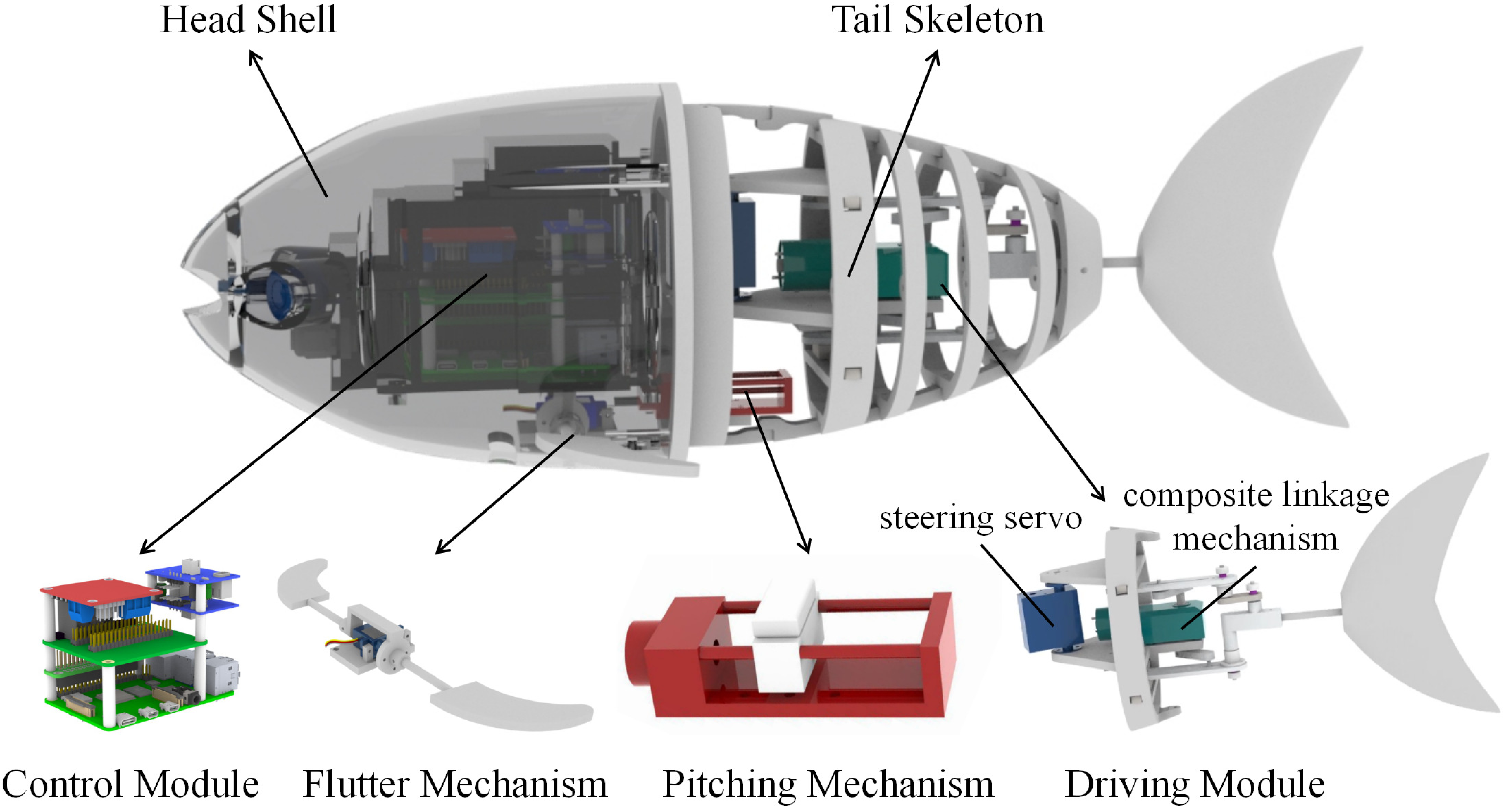}
	\caption{3D model of the modular-designed robotic fish, including head shell, driving module, fishtail skeleton, flutter mechanism, pitching mechanism, and control module.}
	\label{fig9}
\end{figure}
\begin{figure}[t!]
	\centering
	\includegraphics[width=\linewidth]{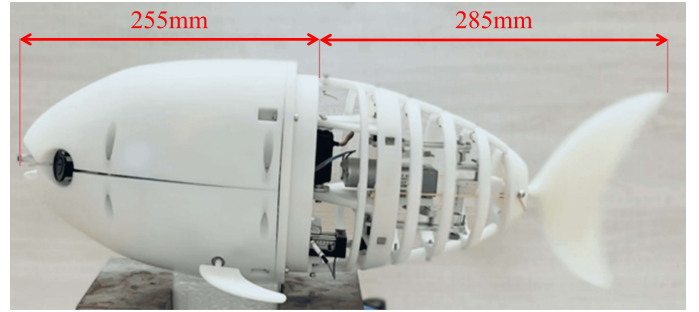}
	\caption{Prototype of the robotic fish created using 3D printing and CNC machining, with dimensions proportionally enlarged to replicate the structural characteristics of live fish.}
	\label{fig10}
\end{figure}
\subsection{Experimental Results and Analysis}
\begin{figure}[b!]
	\centering
	\includegraphics[width=\linewidth]{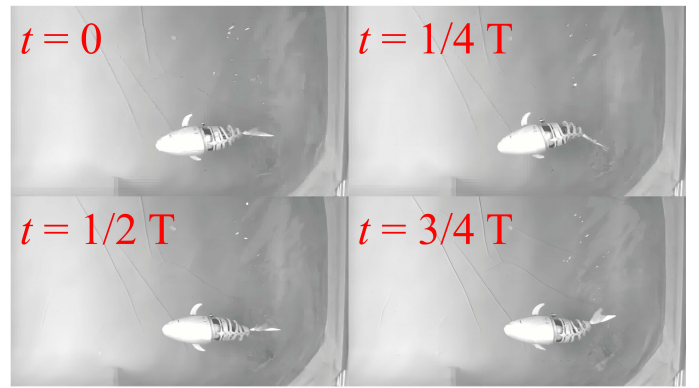}
	\caption{Motion states of the robotic fish at quarter-cycle intervals of a swimming cycle.}
	\label{fig11}
\end{figure} 
	The primary parameters influencing the caudal fin motion, as derived from \eqref{eq10}, include the swing angle amplitude $\theta_{max}$, lateral displacement amplitude $L_{max} = 2L_1$, phase difference $\varphi$, and swing frequency $f$. The subsequent experiments explore the influence of motion parameters, specifically the swing angle amplitude and the swing frequency, on the swimming speed of the robotic fish designed in this study.
\begin{figure}[t!]
	\centering
	\includegraphics[width=\linewidth]{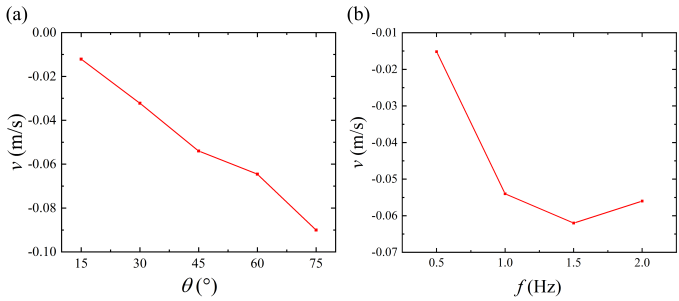}
	\caption{Influence of motion parameters on swimming speed. (a) The relationship between the swing angle amplitude and the steady-state average velocity. (b) The relationship between the swing frequency and the steady-state average velocity.}
	\label{fig12}
\end{figure} 
\subsubsection{Effect of Swing Angle Amplitude on Velocity}
	Fig.~\ref{fig12} (a) illustrates the relationship between the swing angle amplitude and the average speed of the robotic fish in a steady state. The lateral displacement amplitude, phase difference, and swing frequency are maintained at \SI{0.02}{\metre}, $90^{\circ}$, and \SI{1}{\hertz}, respectively. Negative values indicate motion in the direction from the tail to the head of the fish. The experimental results demonstrate an increase in speeds as the swing angle amplitude rises, with the maximum speed of \SI{0.09}{\metre/\second} attained at a $75^{\circ}$ amplitude.
\subsubsection{Effect of Swing Frequency on Velocity}
	Fig.~\ref{fig12} (b) illustrates the relationship between the swing frequency and the average speed of the robotic fish in a steady state. The lateral displacement amplitude, swing angle amplitude, and phase difference are maintained at \SI{0.02}{\metre}, $45^{\circ}$, and $90^{\circ}$, respectively. Due to factors such as limitations in motor power and machining errors, the prototype can achieve a maximum caudal fin swing frequency of \SI{2}{\hertz}. From the experimental results in the graph, it can be observed that as the frequency increases from \SI{0.5}{\hertz} to \SI{2}{\hertz} at intervals of \SI{0.5}{\hertz}, the swimming speed of the robotic fish initially increases and then decreases. The maximum swimming speed reaches around 0.065 m/s at a frequency of approximately 1.5 Hz.
\section{Conclusion}
	This study presents a biomimetic robotic fish propelled by a double-joint mechanism using a composite linkage. The design incorporates a specialized linkage-slider mechanism driven by a single motor, allowing the coupled motion of caudal peduncle translation and caudal fin swing. A comprehensive kinematic analysis is conducted, resulting in equations governing the tail fin's locomotion. Experiments are performed to validate the feasibility of the propulsion mechanism and to explore the influence of motion parameters, including swing angle amplitude and swing frequency, on the swimming speed of the robotic fish. This research paves the way for optimizing bionic fishtail driving devices and delving into the intricate mechanics of fish locomotion.

\bibliographystyle{ieeetr}
\bibliography{IEEEabrv,ref}		
\end{document}